\documentclass{article}

\usepackage[preprint]{neurips_2023}

\usepackage[utf8]{inputenc} 
\usepackage[T1]{fontenc}    
\usepackage{hyperref}       
\usepackage{url}            
\usepackage{booktabs}       
\usepackage{amsfonts}       
\usepackage{nicefrac}       
\usepackage{microtype}      
\usepackage{xcolor}         
\usepackage{nicefrac}		
\usepackage{microtype}		
\usepackage{float}			

\usepackage{subfigure}
\usepackage{algorithm}
\usepackage{algorithmic}
\usepackage{lipsum}		
\usepackage{amsmath, amsthm, amssymb, amsfonts}
\usepackage{newfloat}
\usepackage{sidecap}
\usepackage{graphicx}
\usepackage{textpos} 

\DeclareFloatingEnvironment[name={Supplementary Figure}]{suppfigure}

\DeclareMathOperator*{\argmin}{arg\,min}

\hypersetup{
pdftitle={A multi-criteria approach for selecting an explanation from the set of counterfactuals produced by an ensemble of explainers},
pdfsubject={Machine learning},
pdfauthor={Ignacy St\k{e}pka},
pdfkeywords={Counterfactual Explanations, Explainable AI, Model-agnostic, Machine Learning},
}

\title{A multi-criteria approach for selecting an explanation from the set of counterfactuals produced by an ensemble of explainers}

\author{%
  Ignacy St\k{e}pka\textsuperscript{\rmfamily 1}\thanks{\texttt{ignacy[.]stepka@put.poznan.pl}} \quad Mateusz Lango\textsuperscript{\rmfamily 1,2} \quad Jerzy Stefanowski\textsuperscript{\rmfamily 1} \\
  \textsuperscript{\rmfamily 1}Poznan University of Technology, Poznan, Poland \\
  \textsuperscript{\rmfamily 2}Charles University, Prague, Czech Republic \\
}

\begin{document}

\maketitle

\begin{textblock*}{9cm}(2.6cm,17cm) 
    \centering
    \footnotesize \textcolor{gray}{This is an extended version of the paper that has been published in the International Journal of Applied Mathematics and Computer Science.}
\end{textblock*}

\begin{abstract}

Counterfactuals are widely used to explain ML model predictions by providing alternative scenarios for obtaining the more desired predictions. They can be generated by a variety of methods that optimize different, sometimes conflicting, quality measures and produce quite different solutions. However, choosing the most appropriate explanation method and one of the generated counterfactuals is not an easy task.  Instead of forcing the user to test many different explanation methods and analysing conflicting solutions, in this paper, we propose to use a multi-stage ensemble approach that will select single counterfactual based on the multiple-criteria analysis.  It offers a compromise solution that scores well on several popular quality measures. This approach exploits the dominance relation and the ideal point decision aid method, which selects one counterfactual from the Pareto front. The conducted experiments demonstrated that the proposed approach generates fully actionable counterfactuals with attractive compromise values of the considered quality measures.

\end{abstract}

\section{Introduction}
\label{introduction}



Despite incredible progress in machine learning (ML), the wide adoption of its algorithms, especially in critical domains such as finance or medicine, often encounters obstacles related to the lack of their interpretability.  This is due to the fact that the majority of currently used machine learning methods are black-box models that do not provide information about the reasons behind taking a certain decision, nor do they explain the logic of an algorithm leading to it. Therefore, there is a growing research interest in explainable artificial intelligence  methods~\cite{Bodria-pisa} offering  explanations for the predictions of black-box models.

Counterfactual explanations (briefly \textit{counterfactuals}) are one particular type of such explanations that provide information about how feature values of an example should be changed to obtain a desired prediction of the model (change its decision). 
On the one hand, by interacting with the model using counterfactuals, the user can better understand how the system works by exploring the "what would have happened if..." questions.  This approach to building human understanding of machine learning models has some psychological justifications~\cite{MILLER2019}.
On the other hand, a good counterfactual provides a clear recommendation to the user about what changes are needed in order to achieve the desired outcome. 

There are many practical applications for counterfactual explanations, including loan decisions~\cite{Wachter2017}, recruitment processes~\cite{pearl2016causal}, the discovery of chemical compounds with similar structures but different properties~\cite{counterfactuals_chemistry}, analysis of medical diagnosis results~\cite{counterfactuals_medical}, and many others, see e.g. the recent survey~\cite{Guidotti2022}. For example, consider a scenario where an individual submits a purchase offer for a property, initially rejected by a model assessing such proposals. A counterfactual explanation for this situation reveals the minimal adjustments or enhancements the offeror could implement to ensure the acceptance of their offer (e.g., increase the offered price, relax contingencies). Another practical scenario involves a company training a neural network to assist in the recruitment process for a specific job position, automating the shortlisting of resumes. In this context, counterfactual explanations can be employed to verify that the black-box model does not discriminate against candidates who only differ in terms of a sensitive and non-actionable feature. Furthermore, for a candidate facing rejection, a counterfactual explanation can provide valuable insights into the specific qualifications that were lacking compared to the most similar candidates who were shortlisted.



A counterfactual explanation is expected to be similar to the example that the ML model was queried with, but it should change a class prediction. It leads to a situation where one instance can be explained by many different counterfactuals. 
This has led to the introduction of several desired properties, optimization strategies and \textit{quality measures}, that a counterfactual explanation should possess. Such properties include\footnote{See Sec.~\ref{sec:related} for more details and precise definitions}: \textit{proximity} (counterfactual should be as similar as possible to a given instance), \textit{sparsity} (the number of modified features should be low), \textit{actionability} (the counterfactual should not modify immutable features, such as race, or violate monotonic constraints, e.g. decrease one's age) and \textit{discriminative power} (the generated counterfactual examples should be in the region of the feature space dominated by the expected class), and others.
Even though some of these measures are at least related to each other (e.g., such as proximity and sparsity), many of them are not aligned and even contradictory. For example, proximity encourages the generation of a counterfactual that is as close to the decision boundary as possible, whereas the discriminative power favours explanations in dense areas dominated by the other class, which are most likely to be far from the decision boundary. 
Related user studies~\cite{spreitzer2022evaluating,forster2021capturing} show that human users prefer counterfactuals which, on average, score well on various criteria.  

Generating an appropriate counterfactual explanation is, therefore, a quite challenging task that involves finding a trade-off between divergent aspects of explanation quality.
Nevertheless, the vast majority of counterfactual generation methods optimize only one or two measures, usually aggregated by a weighted sum in the optimized loss function, where the weight is a  hyperparameter of the method, see e.g.~\cite{Wachter2017,Chapman-Rounds-FIMAP,mothilal2020-DICE,vanlooveren2020-cfproto}. The choice of a satisfactory weight value, which controls the trade-off between different aspects of explanation quality, is a not trivial task. The only few methods~\cite{rasouli2022care,Dandl20} that consider multiple quality criteria in the process of obtaining counterfactual explanations, generate a large set of explanations, but leave the task of selecting a final one to the user. We argue, that it is not enough to present numerous explanations to the final user due to the choice overload phenomenon~\cite{iyengar-psychological,stefanowski2023}. Studies~\cite{iyengar-psychological,inbar-regret} show that presenting a limited number of alternatives is superior to presenting too many options, when it comes to human ability to analyze these options and make optimal choices. In order to mitigate these issues, we postulate the need for developing approaches that would reduce the number of proposed counterfactuals and support the user in selecting the most attractive solution.


In this paper, instead of proposing yet another method for generating counterfactuals, we claim that the already existing methods should be sufficient to provide a diversified set of explanations. However, the more challenging problem of selecting a suitable, compromise counterfactual while taking multiple quality criteria into account is still open. Inspired by the research on \textit{classifier ensembles}, which achieve better classification performance by exploiting the predictions from a diversified set of base classifiers~\cite{kuncheva2004combining}, we propose to use an \textit{ensemble of multiple base explainers} to provide a richer set of counterfactuals, each of which establishes a certain trade-off between values of different quality measures  (often referred to as criteria).  We also put forward an approach to significantly reduce the number of considered explanations to a smaller and concise set by constructing a Pareto front, i.e.,~a subset of explanations that are not worse than others on at least one criterion. Then we propose to select a final counterfactual from this front by applying the multiple criteria Ideal Point Method~\cite{steuer1986multiple,skulimowski1990applicability} which does not require additional preference elicitation from  the user, and is computationally efficient. 
To sum up, the main contributions of this paper are as follows:
\begin{enumerate}
\item Proposal of a new approach, integrating an ensemble of explainers algorithm with the multiple criteria analysis to select a counterfactual representing a suitable trade-off between the quality measures.
\item Experimental evaluation of the proposed approach, demonstrating that it  provides the best trade-off between different quality measures for a wide range of user preferences for these criteria.
\item Multi-criteria analysis of counterfactuals generated by various methods, which provides insights into the dominance relationship between them.
\end{enumerate}

\section{Related Works} 
\label{sec:related}

Following~\cite{Wachter2017}, a counterfactual explanation is defined as a perturbation of the instance $x$, denoted as $x'$, that results in a different prediction from the same black box model $b$, i.e., $b(x) \neq b(x')$.

\subsection{Counterfactual Explanation Methods}

Numerous methods for generating counterfactual explanations have been already introduced. According to~\cite{Guidotti2022}, they can be divided into four categories based on their methodological paradigms. Instance-based explainers select the most similar examples with a desired class from the dataset (e.g., FACE~\cite{FACE}). Decision tree approaches approximate the behaviour of a black box model with a decision tree and exploit its structure to generate counterfactual explanations (e.g., ~\cite{FeatureTweak}). The next  approaches optimize a certain loss function by adopting specific optimization algorithms (e.g., Wachter~\cite{Wachter2017}, CEM~\cite{Dhurandhar-CEM}, Dice~\cite{mothilal2020-DICE}, Fimap~\cite{Chapman-Rounds-FIMAP}, CFProto~\cite{vanlooveren2020-cfproto}, ActionableRecourse~\cite{Ustun_2019-actionable-recourse}). Heuristic search strategies find counterfactuals through iterative heuristic choices minimizing the chosen cost function (e.g., Cadex~\cite{Moore-cadex}, GrowingSpheres~\cite{laugel2017-growingspheres}). This taxonomy is only one of many, however it exhibits different approaches to the process of obtaining counterfactual explanations. Another thorough review with an alternative categorization can be found in~\cite{Verma2020}.

Most of the papers introducing counterfactual generation methods do not report experiments that compare them to previous methods on data  benchmarks. In general, the field of counterfactual explanations suffers from a lack of comparative studies examining multiple methods, and the  paper~\cite{Guidotti2022} is the rare exception.

\begin{table}[b]
    \centering
    \caption{Quality measures of counterfactual explanations. $x'$ is the counterfactual for instance $x \in X$. $neigh_k$ is the $k$-th nearest neighbour of $x'$ in $X$. }
    \label{tab:measures}
    \begin{tabular}{p{0.33\linewidth} | p{0.6\linewidth}}
    \hline
        Measure & Definition \\ \hline
         $Proximity(x, x')$ & $distance(x, x')$ \\
         $Feasibility(x')$ & $\frac{1}{k}\sum_{i=1}^{k}distance(x', neigh_i)$  \\
         $Sparsity(x, x')$ & number of features changed in $x$ to get $x'$ \\ 
         $Discriminative\_$ $Power(x')$ & $\frac{1}{k}\sum_{i=1}^{k} \mathbf{1}[[ b(x') = b(neigh_i)]]$ \\
        $Instability(x')$ & $dist(x', x_1')$ where $x_1'$ is a counterfactual obtained for example $x_1 \in X$ being the closest neighbour of $x$ \\\hline
    \end{tabular}
\end{table}

\subsection{Most related methods for generating several counterfactual explanations}

So far, only one paper has considered combining multiple methods for the same data~\cite{guidotti2021ensemble}, where the authors studied the committee of counterfactual explainers. In their proposal base explainers may produce several  explanations which are, then, combined (selected up to the required number) by optimizing a simple two criteria distance-driven aggregation function.

The multi-criteria approach to generating a set of counterfactuals has only been explored in a few papers~\cite{Dandl20,rasouli2022care}. They utilize genetic algorithms to generate a large (tens to hundreds of explanations) set of non-dominated counterfactual explanations based on multiple criteria. However, these approaches do not tackle the problem of reducing the choice overload, and the selection of the best counterfactual explanation is left to the final user or decision-maker.  The needs for adapting  the multi-criteria decision analysis  to support such users is also postulated in \cite{stefanowski2023}.

\subsection{Quality Measures for Counterfactuals}
\label{sec:quality_measures}

The research on counterfactual explanations and their psychological analysis show that they should fulfil some expectations that a human decison-maker might have. 
Here we list the most frequently mentioned properties in literature~\cite{Guidotti2022,Wachter2017}: (1) \textit{validity} - a counterfactual $x’$ has to change the prediction, i.e. $b(x) \neq b(x’)$; (2) \textit{sparsity} - a valid counterfactual should change as few features as possible; (3) \textit{proximity} - a counterfactual should be a result of the smallest perturbation, i.e. $x’$ should be as similar to original $x$ as possible. 
It is also postulated that counterfactuals need to be (4) distributional faithful, i.e. they should be located in feature space regions that ensure their \textit{feasibility} / \textit{plausibility} (as some generation methods may produce examples out-of-data  distribution or with unrealistic feature values). 
In order to take into account fairness and real world utility, many works also introduce (5) \textit{actionability} - a counterfactual should not alter any attributes from $x$ that are sensitive and immutable in certain scenarios (e.g., changing race in a loan application setting).  Other proposed properties are discussed in~\cite{Guidotti2022}.

The above-mentioned properties led to defining different quality measures evaluating either a single counterfactual or a set of counterfactuals. For our further experiments, we chose the most commonly used measures in the literature, see  their specific definitions in Tab.~\ref{tab:measures}.  

In the rest of the paper we use the following notation: $x$ is the original instance, classified by the black box model $b$; $x’$ is the generated counterfactual that corresponds to $x$; \textit{distance} - a distance between two examples calculated with a chosen metric; \textit{neigh$_k$} - the $k^{th}$ closest neighbour to the instance $x$ in the training data~$X$.

\section{The proposed multi-criteria approach}

\begin{figure*}[t]
    \centering    
    \includegraphics[width=\textwidth]{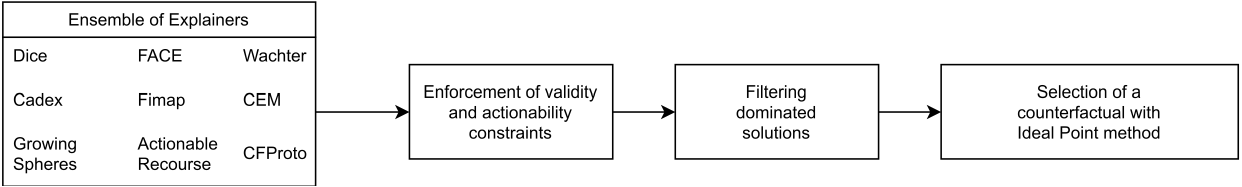}
    \caption{Visualization of the algorithm steps in our proposed approach.}
    \label{fig:flowchart}
\end{figure*}

We propose a new multi-stage approach that integrates an \textit{ensemble of different methods for generating counterfactuals} with a \textit{multi-criteria approach} for selecting the counterfactual that provides a compromise solution with respect to conflicting criteria. It consists of four consecutive steps illustrated in Fig.~\ref{fig:flowchart}. Firstly, each explainer included in the ensemble is queried to generate counterfactuals for a given instance $x$ (Sec.~\ref{sec:ensemble}). Next, all explanations are combined to form a  set of candidate solutions. This resulting set is filtered to remove invalid and non-actionable instances (Sec.~\ref{sec:posthoc-filter}). In the third step (Sec.~\ref{sec:dominance-relation}), we employ the dominance relation to reduce the set of remaining explanations without loss of quality on any considered criterion. As the final step, we use the Ideal Point Method to select the best solution (Sec.~\ref{sec:ideal-point}). 
The pseudocode of the proposed approach can be found in Alg.~\ref{ap:code}. To ease the comprehension of the proposed approach, we present an illustrative example in Sec.~\ref{sec:toy_example}.

\begin{algorithm}[!hb]
    \caption{Pseudocode for the proposed approach}
    \label{ap:code}
    \begin{algorithmic} \vspace{\baselineskip}
        \STATE $b \leftarrow$ black-box classifier
        \STATE $X \leftarrow$ training data
        \STATE $x \leftarrow$ query instance s.t. $x \in X$
        \STATE $E \leftarrow$ set of base explainers
        
        \vspace{\baselineskip}
        STEP 1 (Ensemble of Explainers):
        \STATE $C \leftarrow$ $\emptyset$    
        \FOR{$explainer \in E$}
            \STATE $c \leftarrow explainer(x, X, b)$ \COMMENT{run the base explainer}
            \STATE $C \leftarrow C \cup \{c\}$
        \ENDFOR    
        
        \vspace{\baselineskip}
        STEP 2 (Enforcement of Validity and Actionability):
        \FOR{$c$ in $C$}
            \IF{$\neg is\_valid(c) \lor \neg is\_actionable(c)$}
                \STATE $C \leftarrow C \setminus \{c\}$
            \ENDIF
        \ENDFOR

        \vspace{\baselineskip}
        STEP 3 (Filtering dominated solution):
        \STATE $D \leftarrow \emptyset $\COMMENT{empty set of dominated solutions}
        \FOR{$c$ in $C$}
            \FOR{$d \in (C \setminus \{c\})$}
            \IF{$d$ is better than $c$ on one criterion and better or equal on all criteria }
                \STATE  $D \leftarrow D \cup c$ 
            \ENDIF
            \ENDFOR
        \ENDFOR
\STATE $ND \leftarrow C \setminus D$ \COMMENT{set of non-dominated solutions}

        \vspace{\baselineskip}
        STEP 4 (Selection with Ideal Point)
        \STATE $p \leftarrow vector()$
        \FOR{each criterion $i$}
            \STATE $p_i \leftarrow \max_{c \in ND} c_i $
        \ENDFOR

        \STATE $s \leftarrow \argmin_{c \in ND} distance(c, p)$ 
        
        \vspace{\baselineskip}
        \RETURN $s$
        
    \end{algorithmic}
\end{algorithm}

\subsection{Constructing an ensemble of explainers} 
\label{sec:ensemble}

In order to obtain a set of diversified counterfactuals we construct an ensemble of different methods chosen under the following premises: they are based on different paradigms and thus generate quite diverse explanations, they have positive literature recommendations, and their stable open source implementations are available.  

In Sec.~\ref{sec:experimental-setup}, we list the specific methods used in our ensemble, however we argue, that our approach is general enough to employ different sets of base explainers. While some methods may generate few solutions and others produce a single counterfactual,  experiments have shown that none of the examined methods is superior to others with respect to any of the considered criteria, see~\cite{stepka2023usefulness}. This supports our hypothesis that a set of solutions obtained by using different explanation methods provides a broader and richer set of possible counterfactuals than searching for a single best method. 

\subsection{Enforcement of validity and actionability constraints}
\label{sec:posthoc-filter}

Some of the chosen base explainers may generate counterfactuals which do not change the black-box prediction to the desired value, and also some generated solutions may violate actionability constraints. Therefore, in this step we perform an additional filtering of counterfactuals generated by the ensemble, i.e. we discard explanations which are not \textit{valid} or \textit{non-actionable}. \textit{Actionability} is examined with respect to restricted attributes that have to be distinctly pre-defined for any given dataset.


\subsection{Filtering of dominated solutions}
\label{sec:dominance-relation}

The ensemble of base explainers produces a large number of counterfactuals characterized by different and sometimes conflicting quality criteria (see the discussion of \textit{proximity} and \textit{discriminative power} in Sec.~\ref{introduction}). 
The problem of selecting the best solution is challenging, since the objective comparison of two explanations that excel on different criteria is impossible and largely depends on the preferences of the user/decision-maker. 
Such problems are of interest to the field of multi-criteria decision-aid (MCDA)~\cite{Matthias}, which deals with sets of alternative solutions/decisions $x$ evaluated by many criteria $g_i(x)$. In brief, for each criterion, the direction of the user's preference is defined as increasing (gain criteria) or decreasing (cost criteria).
The values of criteria for considered alternatives may be contradictory, but some of them are more preferred than others because of the preference directions. MCDA methods can solve trade-off between them. 

This leads us to exploiting the \textit{dominance relation}~\cite{Matthias} which is defined as follows.
Let's assume the gain direction of preferences for all criteria $g(x)$ and consider two alternatives - counterfactuals  $x'$ and $y'$. We say that $x'$ \textit{dominates} $y'$, if for each criterion $i$ holds $g_i(x') \geq g_i(y')$, and  exists at least one $i$ for which $g_i(x') > g_i(y')$. In other words, criteria values of $x'$  are better or equal than the corresponding values of $y'$. The dominated counterfactuals can be removed from the set of solutions as they are objectively worse than the \textit{non-dominated} alternatives. %

Then, our approach  constructs \textit{Pareto front} i.e.~it builds a set of all non-dominated counterfactuals obtained by applying the dominance relation on all explanations in a given set. In other words, each counterfactual explanation is examined on whether there exists any other counterfactual that has better or equal scores on all considered criteria. Only if such examples do not exist, the counterfactual is non-dominated and therefore included in the \textit{Pareto front}. Our experiments show that thanks to exploiting this relation, the size of the candidate solution set can be significantly reduced, on average by approx. 80\% (see experiments in Sec.~\ref{sec:analysis-pareto}).

\subsection{Selection of a counterfactual with Ideal Point method}
\label{sec:ideal-point}

As the final set of non-dominated alternatives on the Pareto front may still be too large for a user to analyse manually, it necessary to support the choice of the counterfactual that represents the trade-off best suited to the user. 
There exist many MCDA methods that allow this by acquiring the global model of user preferences for these criteria by interacting with them~\cite{Matthias}.  However, we will follow a simpler approach which is, in our opinion, more suited for our problem and better for carrying out our automatic experiments.


Note that the criteria that characterize different counterfactuals are quality measures (e.g., \textit{proximity}). A typical decision situation considered in MCDA assumes that the user/decision maker is able to compare the alternatives basing on the criteria values (i.e. in the feature space). However, in our case quality measures of counterfactuals are not so intuitive for non-expert humans to interpret and compare.
This also limits the possibility of using typical interactive MCDA methods to elicit the decision maker's preferences, which require accurate assessments about the relative importance of criteria or comparing different variants usually in the original features.




Therefore,  we propose to use the \textit{Ideal Point Method}, which is a simple and compute-efficient approach recommended in the literature for the case of equally important criteria  \cite{branke2008multiobjective,skulimowski1990applicability}.  We briefly explain this method below.

\begin{itemize}
\item Let $D$ be a set of non-dominated counterfactuals  in the Pareto front with $c$ criteria. 
 The ideal solution is artificially created in the criteria space by selecting the best possible value for every criterion.
Assuming that all criteria $g_i(x)$ have an increasing direction of preference (gain), the ideal point $z=[z_1,z_2,..,z_c]$ can be formally defined as having $z_i = \max \{ g_i(x) : x \in D \}$ for all $i =1, ..., c$. Usually $z$ is an abstract point which do not belong to $D$. 
\item Then, for each counterfactual $x \in D$, its distance measure to the ideal point $z$ is computed.
\item The closest counterfactual to $z$ is selected as the final best solution.
\end{itemize}

The Ideal Point method originates from the works on multi-objective mathematical programming, proposed in the previous century~\cite{steuer1986multiple}\footnote{It can be easily extended to use other ways of calculating distances with the hyper-plane connecting the ideal point with the anti-ideal / nadir point \cite{ehrgott2003} (which we will consider in Sect. \ref{impactanalysis})}. It has been further extended for different scalarized distances with criteria weights or to so-called reference points, see discussions in \cite{Matthias,skulimowski1990applicability}. Nevertheless, in our paper we want to show that even the simplest version is sufficient to demonstrate the usefulness of MCDA approach in the proposal of the ensemble of explainers.

\subsection{Approach walkthrough with a toy example}
\label{sec:toy_example}

In order to facilitate the understanding of the proposed method, we will go step by step through the computations for the following instance taken from the Adult\footnote{see Sec.~\ref{sec:experimental-setup} for dataset details} dataset: 
x = \{\textit{age}: 24, \textit{education.num}: 10, \textit{capital.gain}: 0, \textit{capital.loss}: 0, \textit{hours.per.week}: 30, \textit{workclass}: Self-emp-not-inc, \textit{marital.status}: Never-married, \textit{occupation}: Prof-specialty, \textit{race}: Asian-Pac-Islander, \textit{sex}: Male, \textit{native.country}: United-States, \textit{income} $>$50K \}.

First, we run all the explainers from the ensemble to construct counterfactuals. 
Assuming the set of methods used in our experiments (see experimental setup in section \ref{sec:experimental-setup}), 82 counterfactuals are returned. We provide the full list of returned counterfactuals and further details in the online appendix\footnote{\url{https://www.cs.put.poznan.pl/mlango/publications/amcs24.pdf}}. 

We then feed these counterfactuals into the next step of our approach, which focuses on enforcing validity and actionability. The validity filter eliminates counterfactuals that do not change the predicted class from~$>50K$~to~$\leq50K$, reducing the candidate set by 5. Subsequently, actionability enforcement excludes all the alternatives that alter the values of non-mutable features, in this case: \textit{race}, \textit{sex}, and \textit{native.country}. This actionability test removes 18 additional counterfactuals, reducing the total number of candidates to 59.  Nine of the removed counterfactuals suggested changing the person's native country from the United States to other countries such as France, the Philippines, or China. All but one of the non-actionable counterfactuals suggested changing the person's race, and eight of them suggested changing both race and country of birth. This demonstrates the need for actionability and validity testing.

\begin{figure}[t]
  \centering
  \subfigure{\includegraphics[width=0.6\textwidth]{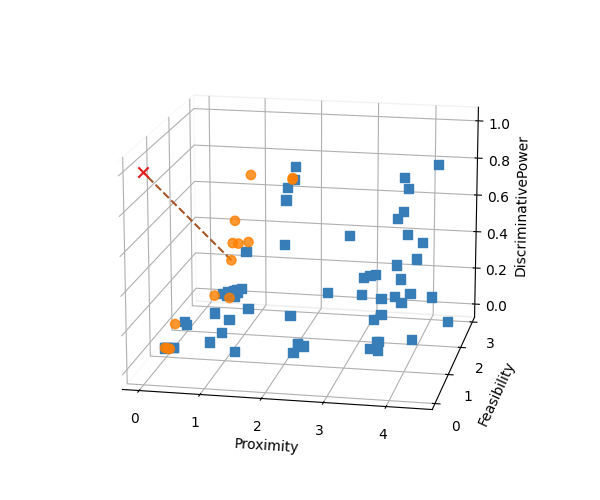}}
  \caption{An example demonstrating the application of dominance relations and the Ideal Point method. Note the optimization directions for \textit{Proximity} and \textit{Feasibility} are min, and for \textit{DiscriminativePower} - max.}
  \label{fig:toy_example}
\end{figure}

In the next step, the multi-criteria analysis of the remaining 59 counterfactuals is already in progress. Initially, we apply the dominance relation to reduce the candidate set to the Pareto front, which consists of 13 alternatives, discarding 46 dominated alternatives. This means that we remove all counterfactuals with criteria values objectively worse on all criteria than a set of other counterfactuals. We visualize this process in Fig.~\ref{fig:toy_example}, where dominated alternatives are denoted with blue squares, and alternatives forming the Pareto front are marked with orange dots.

Subsequently, we calculate the coordinates of the Ideal Point (marked with a red 'x' in the figure) from the Pareto front. The coordinates of the Ideal Point corresponds to the best criteria values obtained by any of the considered alternatives. In this case, these values are 0.04 for Proximity, 0.11 for Feasibility, and 1 for Discriminative Power.

The final step involves calculating the distance between the alternatives from the Pareto front and the Ideal Point, and selecting the closest one (indicated by the dotted red line) as the final counterfactual. The distance is calculated using the Euclidean metric, preceded by feature min-max normalization. In analysed example, the selected counterfactual is:
c = \{\textit{age}: 24.0, \textit{education.num}: 10.0, \textit{capital.gain}: 17327.0, \textit{capital.loss}: 0.0, \textit{hours.per.week}: 30.0, \textit{workclass}: Self-emp-not-inc, \textit{marital.status}: Never-married, \textit{occupation}: Prof-specialty, \textit{race}: Asian-Pac-Islander, \textit{sex}: Male, \textit{native.country}: United-States\}, and it has scores of 0.173 for Proximity, 0.962 for Feasibility, and 0.11 for Discriminative Power. 

\begin{table*}[h]
    \label{tab:datasets-statistics}
    \caption{Characteristics of datasets. From left: \textit{size} - total number of instances, \textit{test size} - the size of holdout test set, \textit{continuous/categorical} - number of continuous/categorical features,  \textit{immutable} - names of features which were designated as non-actionable.}
    \centering
    \begin{tabular}{l|rrrrl}
    \hline
        Dataset & size & test size & continuous & categorical & immutable \\  \hline
         Adult & 32561 &  250 & 5 & 6 & race, sex, native-country\\
         German & 1000 & 100 & 7 & 13 & foreign-worker\\ 
         Compas & 7214 & 250 & 7 & 3 & age, sex, race, charge-degree\\ 
         Fico & 10459 & 250 & 23 & 0 & external-risk-estimate\\\hline
    \end{tabular}
\end{table*}

\section{Experimental evaluation}
\label{experiments}

The aims of experiments are to assess the utility of the proposed approach and to compare the quality of  counterfactuals obtained by different methods.

Firstly, we analyse Pareto front and utility of different steps in our approach. To investigate if constructing an ensemble is justified, we examine the impact, that each step has on the size of a set of candidate counterfactual explanations, as well as the contribution of each of the methods incorporated into the ensemble (Sec.~\ref{sec:analysis-pareto}).

Then, we analyse the results obtained by different explanation methods and our ensemble, by comparing them on different quality metrics (Sec.~\ref{sec:evaluating_measures}).

We  also study the impact of individual components of our approach on the final result (Sec \ref{impactanalysis}).

Finally, using a utility function model of user preferences, we verify whether returned counterfactuals represent useful trade-offs between considered quality criteria (Sec.~\ref{sec:theoretical-user}).

In order to ensure reproducibility and allow future benchmarking we publicly release the code used for experiments\footnote{\url{https://github.com/istepka/MCSECE}}.

\subsection{Experimental setup}
\label{sec:experimental-setup}

\begin{table*}[th]
    \centering
        \caption{Comparison of the explainers used in the ensemble according to their average number of generated counterfactuals per instance per dataset. The columns, listed from left to right, indicate the number of counterfactuals left after applying consecutive steps from our approach: \textit{all} (no filters applied), \textit{val} (after applying the validity requirement), \textit{act} (after applying both the validity and actionability requirements), \textit{front} (after exploiting the dominance relation), and \textit{ideal} (after using the Ideal Point method for selection). The best results are bolded, the second best are underlined and the third best are in italics}
    \label{tab:appendix-pprai}

    \begin{tabular}{l|lllll|lllll}
    \hline
    {Dataset} & \multicolumn{5}{c|}{{Adult}}  & \multicolumn{5}{c}{{German}}  \\ 
    {}  &    all &    val &    act &  front             & ideal             &   all &    val &    act &  front              & ideal             \\ \hline
Dice    &  20.00 &  20.00 &  20.00 &   0.52             & \textit{0.16}     &  20.00 &  20.00 &  20.00 &   0.94 &  \underline{0.33}             \\        
FACE    &  10.00 &   9.68 &   8.16 &   \textbf{3.48}    & \textbf{0.69}     &  10.00 &   9.94 &   0.37 &   \textbf{4.30} &  0.01                \\       
Cadex   &   6.62 &   6.47 &   6.47 &   \textit{1.18}    & \underline{0.34}  &  13.99 &  11.50 &  11.50 &   \underline{2.64} &  \textit{0.28}    \\       
Fimap   &   6.00 &   4.70 &   4.70 &   0.37             & 0.07              &   6.00 &   5.02 &   3.04 &   0.57 &  0.09                         \\
Wachter &   9.60 &   5.03 &   5.03 &   \underline{1.54} & 0.02              &  10.00 &   7.15 &   7.15 &   1.40 &  0.14                         \\
CEM     &   1.00 &   0.66 &   0.66 &   0.26             & 0.00              &   1.00 &   1.00 &   1.00 &   0.09 &  \textbf{0.43}                \\
CFProto &   7.63 &   7.57 &   0.78 &   0.06             & 0.01              &   5.63 &   3.98 &   2.25 &   \textit{1.48} &  0.08                \\
GrowingSpheres &  20.00 &  15.54 &  15.54 &   0.14      & 0.01              &  20.00 &  10.54 &  10.54 &   1.00 &  0.02                         \\
ActionableRecourse  &   0.40 &   0.10 &   0.10 &   0.20  & 0.01             &   0.00 &   0.00 &   0.00 &   0.00 &  0.00                         \\\hline
    \end{tabular}

     \vspace{0.5cm}
     
    \begin{tabular}{l|lllll|lllll}
    \hline
    {Dataset} & \multicolumn{5}{c|}{{Compas}}  & \multicolumn{5}{c}{{Fico}}  \\ 
    {}              &    all &    val &    act &  front & ideal             &   all &    val &    act &  front              & ideal             \\ \hline
Dice                & 20.00 &  20.00 &  20.00 &   \textbf{1.83}  & \textbf{0.35}     &  20.00 &  20.00 &  20.00 &   \underline{2.77} &  \textbf{0.33}             \\ 
FACE                & 10.00 &   9.93 &   0.45 &   0.42  & 0.14              &  10.00 &   9.94 &   0.37 &   0.37 &  0.01                \\ 
Cadex               &  6.00 &   4.78 &   4.74 &   1.08  & \textit{0.19}     &  13.99 &  11.50 &  11.50 &   \textbf{3.72} &  \underline{0.28}    \\ 
Fimap               &  6.00 &   5.58 &   3.30 &   \textit{1.26}  & \underline{0.33}  &   6.00 &   5.02 &   3.04 &   1.13 &  \textit{0.09 }                        \\ 
Wachter             & 10.00 &   6.07 &   6.07 &   1.03  & 0.03              &  10.00 &   7.15 &   7.15 &   \textit{1.65} &  0.14                         \\ 
CEM                 &  1.00 &   1.00 &   1.00 &   0.33  & 0.07              &   1.00 &   1.00 &   1.00 &   0.30 &  0.04                \\ 
CFProto             &  5.75 &   3.65 &   0.73 &   0.13  & 0.00              &   5.63 &   3.98 &   2.25 &   0.95 &  0.08                \\ 
GrowingSpheres      & 20.00 &   8.80 &   8.80 &   \underline{1.73}  & 0.06              &  20.00 &  10.54 &  10.54 &   0.71 &  0.02                         \\ 
ActionableRecourse  &  0.00 &   0.00 &   0.00 &   0.00  & 0.00              &   0.00 &   0.00 &   0.00 &   0.00 &  0.00                         \\  \hline
    \end{tabular}

\end{table*}

Experiments were conducted on four widely adopted datasets in the related literature, namely: Adult, German, Compas and Fico, which characteristics can be found in Tab.~\ref{tab:datasets-statistics}.
For every dataset, we defined a set of immutable attributes  which were later used to evaluate the \textit{actionability} of provided explanations. 
The information about immutability of certain attributes was passed as an additional input to all methods capable of handling it. Note that these subsets could be different depending on the perspective of a particular stakeholder. 

As a black-box classification method, we employed an artificial neural network consisting of two hidden layers with 16-128 neurons in each layer. The number of neurons was optimized for each dataset separately to maximize accuracy on the validation set. We used ReLU activation and dropout between layers. More details of its topology and hyperparameters are provided in the electronic appendix.

To compute the measures described in Sec.~\ref{sec:quality_measures}, it is necessary to select a distance function. We chose the HEOM distance~\cite{Wilson97} due to its  ability to handle both nominal and continuous variables than other measures.   

The evaluated ensemble uses the collection of 9 popular and diversified counterfactual generation algorithms as base explainers, namely Wachter~\cite{Wachter2017}, CEM~\cite{Dhurandhar-CEM}, Dice~\cite{mothilal2020-DICE}, Fimap~\cite{Chapman-Rounds-FIMAP}, Cadex~\cite{Moore-cadex},  FACE~\cite{FACE}, CFProto~\cite{vanlooveren2020-cfproto}, ActionableRecourse~\cite{Ustun_2019-actionable-recourse} and GrowingSpheres~\cite{laugel2017-growingspheres}.  In addition to diversity, we were also guided by their popularity in related works and the availability of their implementations. We used the following open source libraries: CARLA~\cite{Carla}, ALIBI~\cite{ALIBI}, CFEC~\cite{pprai}. 
The methods were mostly used with default parameters in their  implementations, however, to construct a more extensive set of possible explanations, we exploited the possibilities of generating multiple counterfactuals from these methods. 

To obtain several counterfactuals from some of the methods, we apply different strategies, because only Dice natively supports generating a set of explanations (in our case $k=20$). 

The first adopted strategy involves random sampling of counterfactuals discovered during the optimization process, but not selected as the final solution by the method. Therefore, it not only gathers the explanation that the method ultimately selects, but also incorporates randomly sampled counterfactuals discovered during optimization that may perform better on some criteria that the method does not directly optimize. We apply this strategy to CFProto ($k=10$) and Wachter ($k=10$).

The second strategy is to restart the method with different set of hyperparameters to slightly change the optimization process and obtain different explanation. We use this strategy for GrowingSpheres ($k=20$ restarts with different random seeds), FACE ($k=10$ restarts with different random seeds), Cadex ($k=15$ different numbers of features to change ranging from 1 to 14), Fimap ($k=7$ different combinations of parameters for the objective function\footnote{
The following combinations of (Gumbel-softmax temperature $\tau$, L1 regularization, L2 regularization) were applied: (0.1, 0.001, 0.01),(0.1, 0.05, 0.5),(0.2, 0.01, 0.1),(0.2, 0.08, 0.8),(0.5, 0.001, 0.01),(0.5, 0.01, 0.5)}).  

While selecting the final counterfactual with our approach, we analysed three criteria: \textit{proximity}, \textit{feasibility} and \textit{discriminative power}. Recall that in MCDA criteria must form a coherent family of diverse views on the problem. Indeed, these three criteria were relatively poorly correlated in our preliminary experiments~\cite{stepka2023usefulness}.
As a distance function for the Ideal Point method, we considered Manhattan distance ($L_1$), Euclidean distance ($L_2$) and Chebyshev distance($L_\infty$) (following literature such as \cite{branke2008multiobjective,skulimowski1990applicability}).

\begin{table*}[ht]
\caption{The results obtained for German dataset. The best results are bolded, the second best are underlined and the third best are in italics.}
\label{tab-german}
    \centering
\begin{tabular}{l|rrrrrrr|r}
\hline
Method & prox $\downarrow$& feas $\downarrow$& dpow $\uparrow$& spars $\downarrow$& instab $\downarrow$& cover $\uparrow$& act $\uparrow$ & rank $\downarrow$ \\ \hline
Dice  &       1.69 &             3.92 &                    0.44 &      \underline{1.93} &         4.15 &      \textbf{1.00} &        \textbf{1.00} &  3.14 \\
FACE                     & 5.05          & \textbf{1.91} & 0.60          & 8.12          & 3.82          & \textbf{1.00} & 0.98          & 3.14 \\ 
Cadex                    & \textit{1.38} & 3.74          & 0.41          & 2.64 & 3.87          & 0.97          & 0.97          & 5.43 \\
Fimap                    & 6.85          & 3.01          & 0.60          & 9.91          & 3.71          & 0.97          & 0.97          & 4.57 \\
Wachter    &      11.67 &             7.29 &                    0.64 &     14.65 &         5.91 &      0.37 &        0.37 &  5.14 \\
CEM                      & \textbf{0.62} & 4.18          & 0.31          & \textit{2.15}    & 3.99          & 0.13          & 0.13          & 7.00 \\
CFProto  &       3.56 &             4.40 &                    0.48 &      4.79 &         4.53 &      0.99 &        0.91 &  5.29 \\
GrowingSpheres           & 7.65          & 5.79          & 0.60          & 10.73         & 5.42          & \textbf{1.00} & \textbf{1.00} & 2.71 \\
ActionableRecourse       & \underline{1.01}    & 3.55          & 0.44          & \textbf{1.39} & \underline{3.60}    & 0.23          & 0.23          & 6.29 \\ \hline
random selection          & 4.39          & 3.95          & 0.50          & 6.28          & 4.61          & \textbf{1.00} & 0.98          & 6.86          \\
our approach (Manhattan) & 3.83          & \underline{2.15}    & \textbf{0.85} & 6.06          & \textbf{3.50} & \textbf{1.00} & \textbf{1.00} & \textbf{1.86} \\
our approach (Euclidean) & 3.21          & \textit{2.46} & \underline{0.80}    & 4.99          & \textit{3.68} & \textbf{1.00} & \textbf{1.00} & \underline{2.00} \\
our approach (Chebyshev) & 2.90          & 2.70          & \textit{0.74} & 4.38          & 3.71          & \textbf{1.00} & \textbf{1.00} & \textit{2.14} \\   \hline    
\end{tabular}

\end{table*}

\begin{table*}[p]
\caption{The results obtained for Adult dataset.}
\label{tab-adult}
    \centering
\begin{tabular}{l|rrrrrrr|r}
\hline
Method & prox $\downarrow$& feas $\downarrow$& dpow $\uparrow$& spars $\downarrow$& instab $\downarrow$& cover $\uparrow$& act $\uparrow$ & rank $\downarrow$ \\ \hline
Dice                &       1.03 &             0.77 &                    0.37 &      \underline{1.65} &         1.13 &      \textbf{1.00 }&        \textbf{1.00} &  3.00 \\
FACE                     & 0.98          & 0.90          & 0.36          & \textit{1.92}     & 1.10    & 0.10   & 0.10  &      3.43 \\ 
Cadex                    & \underline{0.20}          & 0.30          & 0.17          & 2.27 & 0.65   & 0.99   & 0.99   &   4.86 \\
Fimap                    & 2.12          & 0.35          & 0.59          & 5.75          & 1.17          & 0.99          & 0.99          & 4.00 \\
Wachter                  & 4.36          & 1.23          & 0.84          & 7.76          & 3.07          & 0.52          & 0.20          &   5.29 \\
CEM                      & \textbf{0.13} & 0.32          & 0.17          & \textbf{1.16} & 0.67          & 0.66          & 0.66          &    6.14 \\
CFProto                  & 1.45          & 1.13          & 0.25          & 7.00          & 2.97          & 0.98          & 0.06          &   6.14 \\
GrowingSpheres           & 2.87          & 1.39          & 0.47          & 6.17          & 1.70          & 0.99          & 0.99          &     4.14 \\
ActionableRecourse       & 1.14          & \textbf{0.10} & 0.70          & 3.72          & \textbf{0.56} & \textbf{1.00} & 0.84          &      6.57 \\ \hline
random selection          & 1.55          & 0.81          & 0.38          & 3.56          & 1.17          & \textbf{1.00} & 0.88          &  3.86 \\
our approach (Manhattan) & 1.02          & \underline{0.17}      & \textbf{0.94} & 3.34   & \underline{0.63}   & \textbf{1.00} & \textbf{1.00} &  \textbf{1.86} \\
our approach (Euclidean) & 0.99          & \textit{0.21} & \underline{0.93}    & 3.22   & \underline{0.63}  & \textbf{1.00} & \textbf{1.00} & \underline{2.00} \\
our approach (Chebyshev) & \textit{0.96} & 0.26    & \textit{0.89} & 2.97  & 0.66    & \textbf{1.00} & \textbf{1.00} &  \textit{2.14} \\\hline
\end{tabular}

\end{table*}

\begin{table*}[p]
\caption{The results obtained for Compas dataset.}
\label{tab-compas}
    \centering
\begin{tabular}{l|rrrrrrr|r}
\hline
Method & prox $\downarrow$& feas $\downarrow$& dpow $\uparrow$& spars $\downarrow$& instab $\downarrow$& cover $\uparrow$& act $\uparrow$ & rank $\downarrow$ \\ \hline
Dice &       0.93 &             0.80 &                    0.34 &      \textit{1.71} &         1.33 &      \textbf{1.00} &        \textbf{1.00} &  3.00 \\
FACE                     & 0.53          & \textbf{0.04} & 0.69          & 3.55          & \textbf{0.15} & \textbf{1.00} & 0.07          & 4.14 \\ 
Cadex                    & \underline{0.29}          & 0.23          & 0.15          & 2.72          & 0.36          & 0.98          & 0.97          & 5.43 \\
Fimap                    & 0.52          & \underline{0.11}          & 0.69          & 3.53          & \textit{0.17} & 0.99          & 0.58          & 5.00 \\
Wachter &       0.69 &            \textit{ 0.12} &                    0.73 &      2.92 &         0.36 &      0.67 &        0.67 &  5.00 \\
CEM                      & 0.33          & 0.32          & 0.29          & \underline{1.57}          & 0.38          & \textbf{1.00} & \textbf{1.00} & 3.14 \\
CFProto &       0.91 &             0.21 &                    0.28 &      3.17 &         0.41 &      0.54 &        0.13 &  6.29 \\
GrowingSpheres           & \underline{0.29}          & 0.19          & 0.18          & 3.09          & 0.32          & 0.80          & 0.80          & 5.57 \\
ActionableRecourse       & \textbf{0.07} & 0.32          & \textbf{0.89} & \textbf{1.00} & 0.66          & 0.00          & 0.00          & 5.29 \\ \hline
random selection          & 0.63          & 0.37          & 0.40          & 2.74          & 0.61          & \textbf{1.00} & 0.70          & 3.86 \\ 
our approach (Manhattan) & 0.54          & \textit{0.12}          & \underline{0.87}          & 2.40          & \underline{0.28}          & \textbf{1.00} & \textbf{1.00} & \textbf{2.00} \\
our approach (Euclidean) & 0.55          & 0.14          & \textit{0.86} & 2.45          & \underline{0.28}          & \textbf{1.00} & \textbf{1.00} & \underline{2.14} \\
our approach (Chebyshev) & 0.55          & 0.17          & 0.84          & 2.44          & 0.30          & \textbf{1.00} & \textbf{1.00} & \textit{2.29} \\ \hline
\end{tabular}

\end{table*}  
\begin{table*}[p]
\caption{The results obtained for Fico dataset.
Note that ActionableRecourse method is missing, as it failed to provide any valid counterfactual.}
\label{tab-fico}
    \centering
\begin{tabular}{l|rrrrrrr|r}
\hline
Method & prox $\downarrow$& feas $\downarrow$& dpow $\uparrow$& spars $\downarrow$& instab $\downarrow$& cover $\uparrow$& act $\uparrow$ & rank $\downarrow$ \\ \hline
Dice  &       1.11 &             2.15 &                    0.38 &      \textbf{1.88} &         2.53 &      \textbf{1.00} &        \textbf{1.00} &  3.00 \\
FACE                     & 2.34          & \textbf{0.82} & \underline{0.70}          & 17.88         & \underline{1.78} & \textbf{1.00} & 0.03          & 3.57 \\  
Cadex                    & 0.92          & 1.72          & 0.38          & 7.66          & 2.04          & \textbf{1.00} & \textbf{1.00} & 3.14 \\
Fimap                    & 1.55          & 1.78          & 0.62          & 16.01         & \textit{1.87}          & 0.98          & 0.62          & 4.71 \\
Wachter  &       6.66 &             3.50 &                    \textbf{0.81} &     18.39 &         6.45 &      0.49 &        0.49 &  4.71 \\
CEM                      & 1.12          & 2.08          & 0.50          & \underline{5.80} & 2.47          & \textbf{1.00} & \textbf{1.00} & 2.71 \\
CFProto  &       \textbf{0.82} &             \underline{1.53} &                    0.35 &     10.82 &         \textbf{1.76} &      0.58 &        0.39 &  6.29 \\
GrowingSpheres           & 1.44          & 1.90          & 0.35          & 16.48         & 2.39          & 0.97          & 0.97          & 5.29 \\ \hline
random selection          & 1.30          & 1.84          & 0.40          & 9.67          & 2.21          & \textbf{1.00} & 0.84          & 3.86 \\
our approach (Manhattan) & \underline{0.87}          & \underline{1.51}          & 0.60          & 7.33 & 1.89 & \textbf{1.00} & \textbf{1.00} & \textit{2.57} \\
our approach (Euclidean) & \textit{0.90} & 1.59          & \textit{0.63} & 7.48          & 1.93          & \textbf{1.00} & \textbf{1.00} & \textbf{2.14}    \\
our approach (Chebyshev) & 0.99          & 1.66          & \textit{0.63} & \textit{7.11}          & 2.03          & \textbf{1.00} & \textbf{1.00} & \textbf{2.14}\\\hline
\end{tabular}

\end{table*}

\subsection{Analysis of the Pareto front of counterfactuals generated by the base explainers in the ensemble}
\label{sec:analysis-pareto}
    
In the first experiment, we take a closer look at the degree of base explainers' contribution to the Pareto front.
For each dataset and base explainer method, we computed the average number of counterfactuals generated for an instance (all), the number of valid  counterfactuals, i.e.~those that change the model's prediction (val), the number of actionable and valid counterfactuals (act), the number of non-dominated, valid, actionable counterfactuals on the Pareto front (front) and  finally the percentage of cases for which the counterfactual generated by a given method was selected as the final answer (ideal).
The results are reported in Tab.~\ref{tab:appendix-pprai} and some additional data visualizations are provided in the appendix.

Results show, that there is no single method that is superior to others considered in the experiments. Some of the methods clearly contribute much more frequently to the Pareto front, but there is no method which contribution is negligible on all examined datasets. Contribution to the Pareto front can be treated as a good indicator of the performance of the method across different measures, as it shows that counterfactuals generated from some methods dominate others on all quality measures. Building on that we justify the utility of incorporating all these methods into the ensemble, as all of them  produce best explanations, but with differing frequencies. 

The further analysis  indicates, that all steps of our approach eliminate many  counterfactuals. First, enforcing \textit{validity} constraints reduces the number of considered candidates by 17\%. Second, examining \textit{actionability} eliminates 16\% of examples from previous step. Third, the use of dominance relation reduces the remaining set by another 83\%. Finally, the Ideal Point Method selects one compromise counterfactual from the remaining set of explanations. 
By employing consecutive steps of or approach, with an exclusion of the last one, it shrinks the original set of explanations, without the loss of quality, by on average 88\%. 

\subsection{Evaluating counterfactuals with different quality measures}
\label{sec:evaluating_measures}

In the second experiment, we compare the performance of the proposed approach with other methods for generating counterfactuals.
The quality of the generated counterfactual explanations is evaluated using a wide spectrum of measures used in the related works: \textit{proximity} (prox), \textit{feasibility} (feas), \textit{discriminative power} (dpow), \textit{sparsity} (spars), \textit{instability} (instab), and \textit{actionability} (act) -- all of them were discussed in Sec.~\ref{sec:quality_measures}. Additionally, to further assess the reliability of the methods in finding counterfactuals, we report the \textit{coverage} (cov) metric, which represents the ratio of instances for which a counterfactual is found.
The results of experiments for German, Adult,  Compas, Fico datasets can be found in Tables~\ref{tab-german}, \ref{tab-adult}, \ref{tab-compas} and \ref{tab-fico}, respectively.

Among the methods under consideration, the proposed approach was the only one that was always able to generate an actionable counterfactual for all instances of tested datasets.
Looking at the remaining five quality aspects (other than \textit{actionability} and \textit{coverage}), the proposed approach rarely obtains the highest score, however, for the vast majority of cases it obtains one of the top three results.
This was expected since our method looks for a trade-off between multiple quality measures, therefore it is not surprising that in the individual ranking for each measure it may not obtain superior results.

We also ranked the data for each quality measure and computed the average rank for each method (the lower, the better) -- the results can be observed in the last column of Tables~\ref{tab-german}-~\ref{tab-fico}.
Taking all the metrics into account, the proposed approach always achieves the best (lowest) rank for all datasets under study.
The lowest rank  for Adult, German and Compas datasets is achieved by the variant of our approach employing Manhattan distance. Only for the Fico dataset the order of three best methods is different. Nevertheless, for all datasets included in the experiments, variants of our approach take consistently best three ranks, meaning that they offer the best compromise between all seven considered quality measures. 

As a form of ensemble baseline, we also report the results of the ensemble that includes all the base explainers of the proposed method, but chooses the final counterfactual at random (random selection) without employing our algorithm.
This ensemble was never better than any of the three tested variants of our approach for any dataset and any quality measure, demonstrating that the performance of our ensemble is a result of choosing an appropriate final counterfactual with multi-criteria analysis (i.e.~applying dominance relation and Ideal Point Method selection) rather than just using an ensemble of different methods.

Regarding the comparison of different variants of our method employing different distance measures, the differences between the obtained scores are usually very small (except for the Sparsity on the German dataset, where the difference between the extreme scores is 1.68). The variant employing Manhattan distance obtains slightly lower average rank for most datasets, therefore we choose this variant to represent our method in the next experiment.


\subsection{Analyzing the impact of the elements in our approach}
\label{impactanalysis}
In this section, we provide a concise analysis of how the components of our approach influence the selection of final counterfactuals and their corresponding evaluation measures. The experimental data used for this analysis is available in the appendix.

The first element in our approach emphasizes validation and actionability enforcement. Our experiments demonstrate the critical role this element plays in achieving fully actionable and valid counterfactuals. Without it, the scores for the \textit{Validity} and \textit{Actionability} criteria range from 24\% to 54\% and 78\% to 98\%, respectively (varies by dataset). Such performance falls short of suitability for most real-world scenarios, proving the usefulness of this operation.

The second element in our approach involves applying the dominance relation to filter out dominated alternatives. Omitting this component does not lead to a decrease in the values of various quality measures. This is because, in our approach, the Ideal Point selects the counterfactual closest to the (0,0,0) point in the \textit{Proximity-Feasibility-(1-DiscriminativePower)\footnote{For simplicity, we invert the DiscriminativePower criterion to have min optimization direction.}}  min-max normalized criteria space, which is attained through min-max normalization performed on the dataset. Consequently, the dominance relation do not enhance the performance of the Ideal Point method. 
Nevertheless, removing dominated, i.e. worse, solutions is necessary for considering other multi-criteria methods. Moreover, as we have already shown in experiments (see Table \ref{tab:appendix-pprai}), this step significantly reduced the number of solutions. 

The final component of our approach involves selecting the counterfactual by means of the Ideal Point method. On average, this method yields significantly better-scoring counterfactuals compared to random selection from the Pareto front. An interesting comparison lies in contrasting the Ideal Point method with the simple unweighted sum ($s = \text{Proximity} + \text{Feasibility} + (1-\text{Discriminative Power})$) of the criteria scores. Our experiments (see the appendix) reveal that the variant employing Manhattan distance achieves the best average scores across the examined datasets. This variant is equivalent to the unweighted sum ($s = c_{\text{Manhattan}}$), as, in the previously mentioned min-max normalized criteria space, the Ideal Point coordinates are $p = (0,0,0)$. Therefore, the distance to the Ideal Point $c_{\text{Manhattan}} = |p_0-\text{Proximity} + p_1-\text{Feasibility} + p_2-(1-\text{Discriminative Power})|$ is just a summation of scores on all criteria, equivalent to the unweighted sum: $s = c_{\text{Manhattan}}$. It is worth noting that other variants of our approach using Euclidean or Chebyshev distance are not equivalent and select different alternatives. 

While this paper primarily utilizes the simplest multi-criteria selection method, we also conducted preliminary experiments (included in the appendix) with a slightly more advanced selection method based on distance to the nadir-ideal plane \cite{ehrgott2003}, showing slightly superior results compared to the unweighted sum selection. This underscores the value of employing multi-criteria selection methods. Even this simple distance methods are also useful for possible further extensions of the methods towards interactive dialogue with the decision-maker.

\begin{figure*}[h]
\centering
  \subfigure{\includegraphics[width=0.49\textwidth]{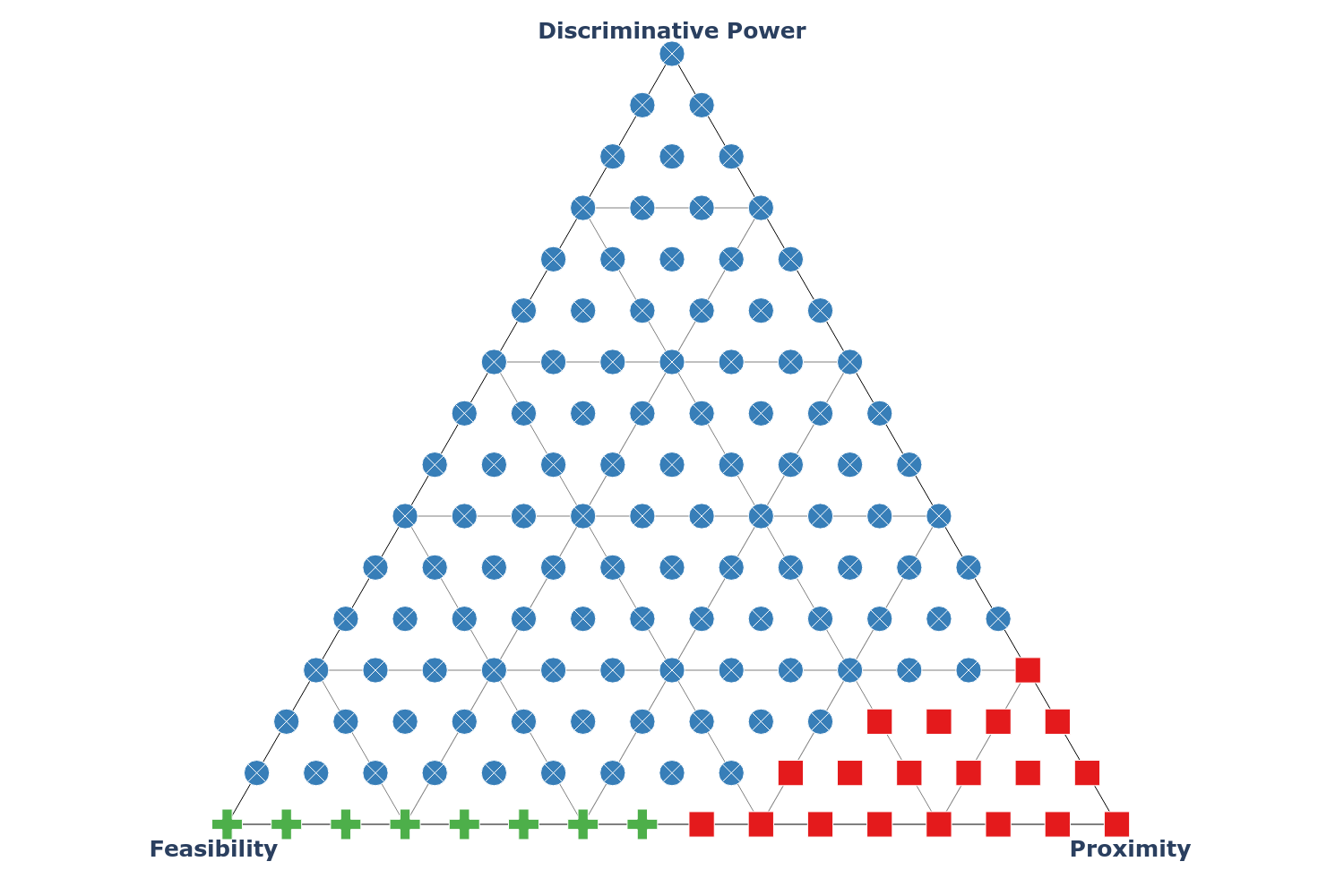}}
  \subfigure{\includegraphics[width=0.5\textwidth]{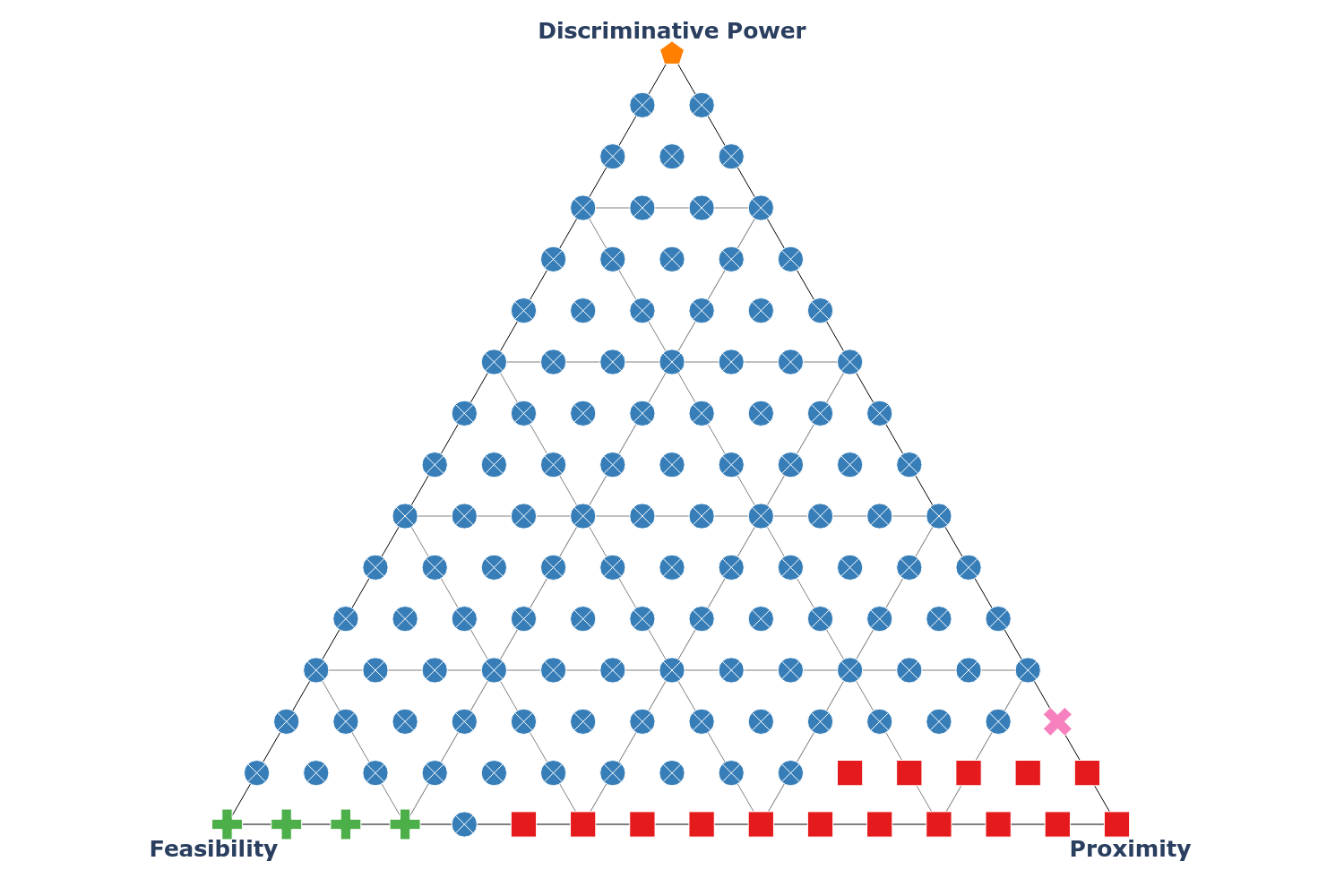}}
  \\
  \subfigure{\includegraphics[width=0.49\textwidth]{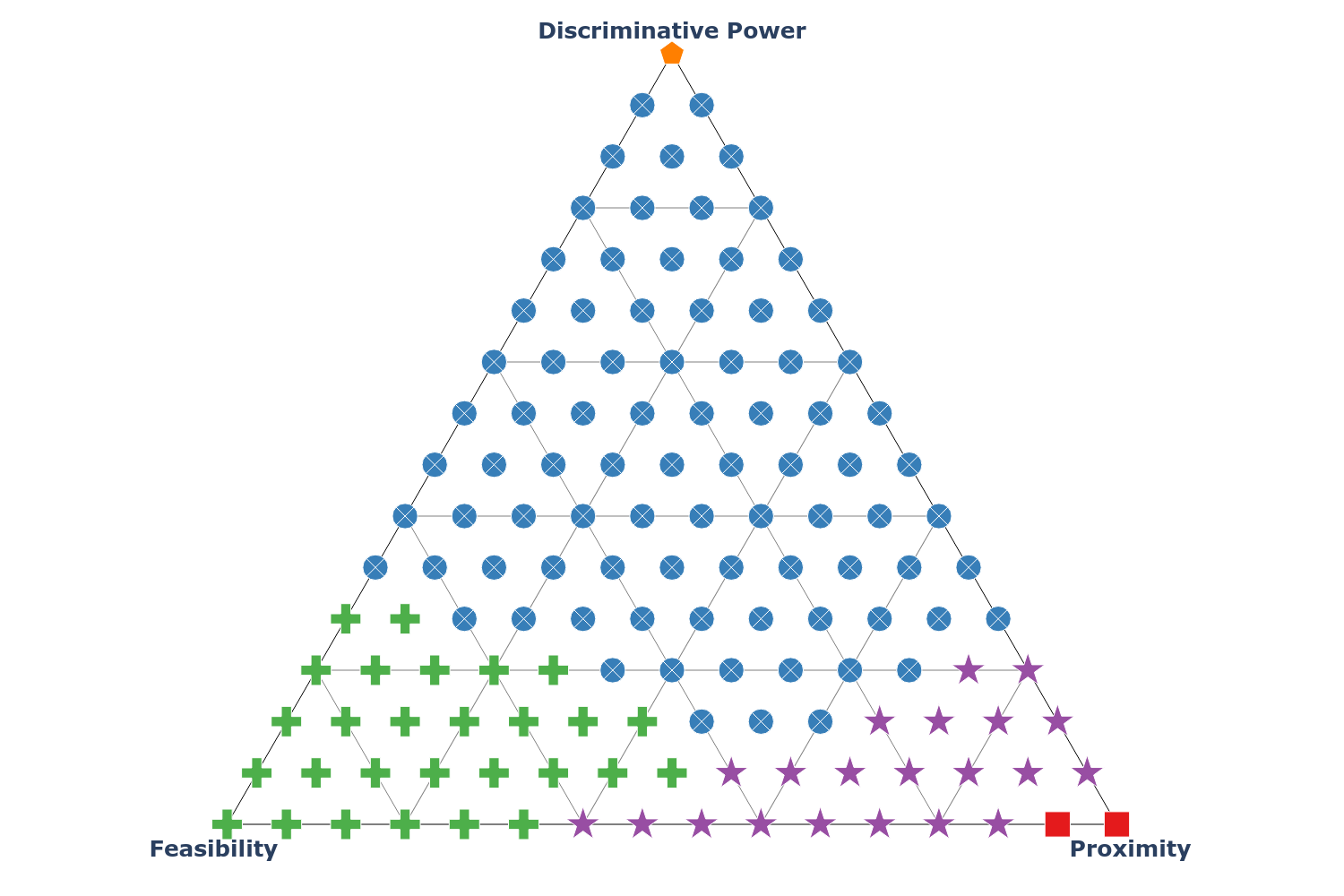}}
  \subfigure{\includegraphics[width=0.5\textwidth]{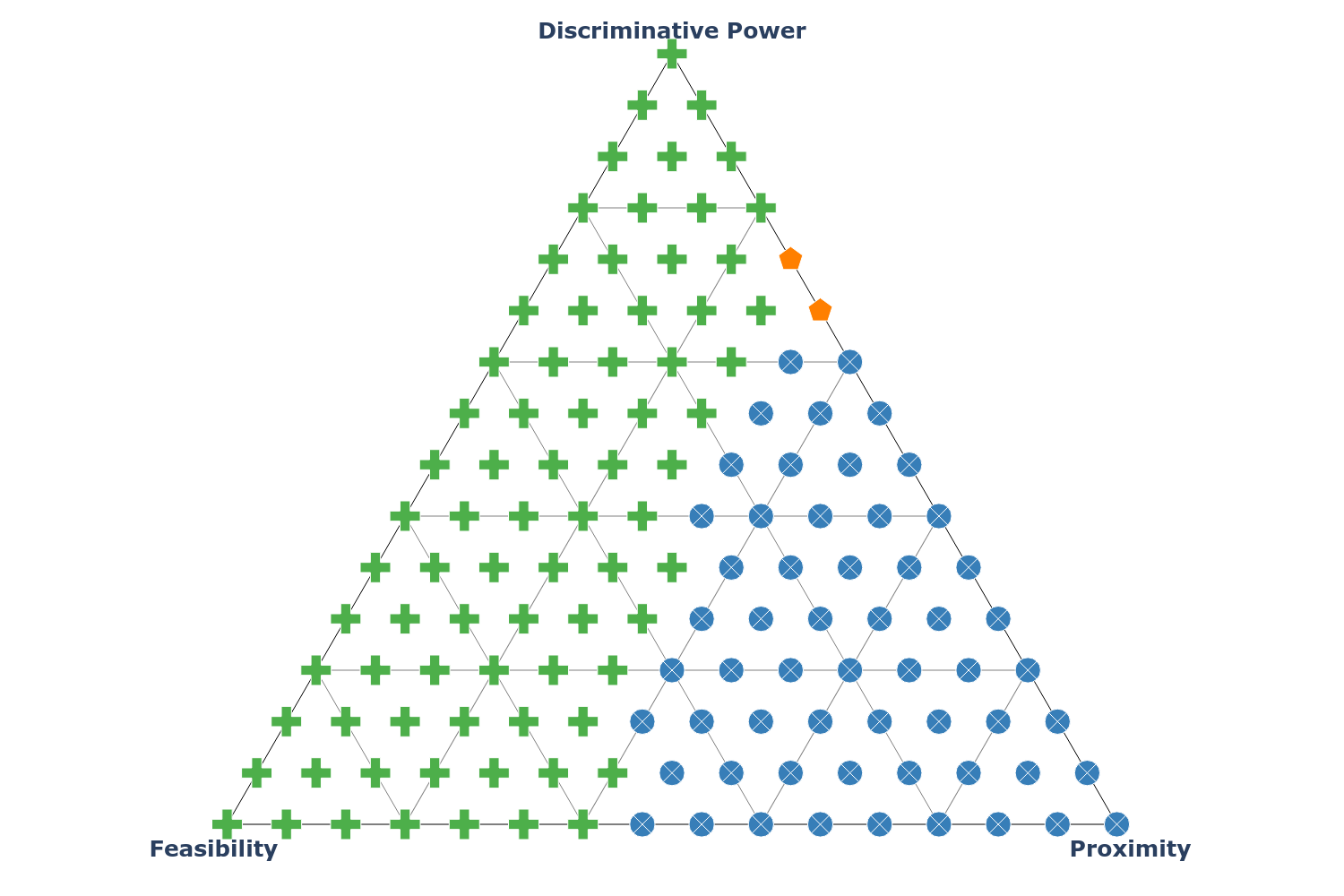}}
  \\
  \subfigure{\includegraphics[width=0.7\textwidth]{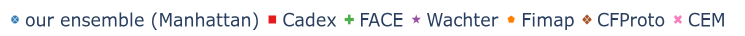}}
      \caption{
    The barycentric plots depicting the best method for a given dataset according to the utility functions with different weights assigned to quality criteria. The datasets are: German (\textit{upper-left}), Adult (\textit{upper-right}), Compas (\textit{lower-left}), Fico (\textit{lower-right}).
      }
  \label{fig:multiplot}
\end{figure*}

\subsection{Using user preference models to evaluate selected trade-offs between different quality measures}
\label{sec:theoretical-user}
The main aim of our work is to generate a counterfactual for a given instance, which represents a suitable trade-off between various aspects of explanation quality. 
Let us recall that in our approach we do not use criterion weights due to their difficulty in estimating by humans. However, for the purpose of examining possible trade-off criteria in potential models of decision-makers' preferences, we decided to experimentally simulate  a simple utility function model, where we analyzed the impact of all possible weight configurations.
Since an objective comparison of different non-dominated trade-offs is impossible without additional information about user preferences towards optimizing particular quality criteria, we employ a simple mathematical model of user preference to verify the utility of counterfactuals under all possible configuration of preferences.
More concretely, we model the utility of an explanation as a weighted sum of three selected quality criteria:
\begin{equation}
\begin{split}
U(x') = w_p\cdot &\textrm{Proximity}(x') \\&+ w_d\cdot \textrm{DiscriminativePower}(x') \\&+ w_f\cdot \textrm{Feasibility}(x')
\end{split}
\end{equation}
where $w_p,w_d,w_f\geq 0$, $w_p+w_d+w_f=1$ are parameters which control the importance of individual quality measures for a potential user. It is assumed that the user should select the solution which maximizes the utility function. 

Since the space of all possible utility functions has only three parameters ($w_p,w_d,w_f$) which sum up to 1, it can be easily visualized on a 2-dimensional plot (see Fig. \ref{fig:multiplot}) with the Barycentric coordinate system.
The barycentric plot takes the form of an equilateral triangle with each vertex associated with one weight of the utility function. 
Each point inside the triangle represents one possible user utility function, i.e.~one possible setup of $w_p,w_d,w_f$ weights.
For instance, a point at the vertex corresponding to \textit{proximity} ($w_p$) represents the utility function with $w_p=1$ and $w_d=w_f=0$. 
A point in the middle of the edge connecting the vertices corresponding to \textit{proximity} ($w_p$) and fidelity ($w_f$) represents the utility function with $w_p=w_f=0.5$ and $w_d=0$. 
Finally, the central point in the middle of the triangle corresponds to a user having equal preferences for all quality criteria ($w_p=w_d=w_f=\frac{1}{3}$).

We computed the utilities of the counterfactuals returned by all the methods under study using utility functions for all\footnote{More precisely, we evaluated a grid of 136 weight combinations, evaluating weight values from 0 to 1 with a step of $\frac{1}{16}$.} possible combinations of weights values.
Later, we verified which counterfactual method would be selected as the preferred one by a potential user with such preferences.
The results of this experiment are visualized on the Fig.~\ref{fig:multiplot}.

For the Adult dataset, despite being uninformed about user preferences and always selecting the same counterfactual with the Ideal Point method, our proposed approach would be selected as the best one according to approximately 85\% of all possible utility functions.
Only when the user has a strong preference towards one criterion, the proposed approach loses to CADEX on \textit{proximity} and to FACE on \textit{feasibility}.
For the user solely interested in \textit{discriminative power} the counterfactuals produced by Fimap would be the most appropriate.
Similar observations can be made for the German and Compas datasets, where the results of our method constitute the best trade-off for all users who are not too strongly biased towards one particular criterion.

Only for the Fico dataset, the counterfactuals produced by FACE seem to represent a good trade-off between Feasibility and Discriminative Power for many cases.
However, it is important to note that for this dataset FACE generates actionable explanations only for 3\% of examples (see Tab.~\ref{tab-fico}).
Therefore, this result does not demonstrate inefficiency of our selection procedure since we automatically discard all non-actionable explanations beforehand.
Nevertheless, our proposed method achieve good trade-offs for users who are more inclined towards obtaining explanations similar to the instance being explained (\textit{proximity}) but  who are also interested in reasonable \textit{feasibility} and \textit{discriminative power}. In other words, unless decision-maker is interested solely in counterfactual explanations that score high only on one criterion, our proposed method provides more preferred counterfactuals.

\section{Conclusions and Future Works}
The main message of our work is to promote the use of multi-criteria decision analysis (MCDA) to select contractual explanations for predictions made by black-box machine learning models. 
The proposed approach has at least two original contributions to current research. Firstly, we propose to construct an ensemble of various explanatory methods that are effective in generating a fairly large (in our experiments, about 80-90) set of diverse solutions. Secondly, we employ further multi-criteria analysis to first, reduce the number of considered counterfactuals to only non-dominated explanations, and thirdly, to support the selection of a solution that offers a compromise between the values of the considered evaluation measures.

In this work, we have consciously chosen relatively simple and well-known  MCDA proposals: (1) the use of the dominance relation - which is the only approach that is fully objective to filter out the worse evaluated solutions; (2) the Ideal Point method - which is also a no preference approach (i.e., it does not require any acquisition of user preferences on the relative importance of criteria -- weights or comparisons of alternatives). It is also computationally simple to run automated experiments.

Additional experiments (Sec \ref{impactanalysis}) also showed that all steps of our approach are essential. Moreover, even a relatively simple multi-criteria ideal point method leads to good choices of counterfactuals and can be easily further extended, e.g. to the ideal-nadir version.

Despite the simplicity of these multi-criteria methods, we believe that the presented experiments demonstrated the utility of our approach. Indeed, the selected counterfactuals are competitive with the solutions offered by the best of the single methods used, which often optimize only one criterion. Furthermore, our additional experiments simulating the utility model of a potential decision maker show that when they do not have a strong preference for a single criterion, the proposed multi-criteria approach is highly beneficial and provides a good trade-off between criteria.

Furthermore, our comparative study reveals that no method is superior to others in all criteria. Therefore, we conclude that more attention should be given to comparing different counterfactual generation methods using multi-criteria analysis to highlight the various trade-offs made by these methods.

We also argue, that looking at explanations from different perspectives should be studied more extensively, and we believe that multi-criteria analysis is the natural choice for this type of investigation. As a next step, we suggest that research should focus on whether the proposed framework aligns with the human perspective, and whether humans make trade-offs between criteria when selecting the best explanations or whether they prefer only one of them. Moreover, it is worth considering adaptations of interactive, dialogue multi-criteria selection methods, which will be the subject of further research.

\begin{ack}
The research has been partially supported by 0311/SBAD/0743,  0311/SBAD/0740 PUT University grants and TAILOR, a project funded by EU Horizon 2020 research and innovation programme under GA no. 952215.

\end{ack}

\bibliographystyle{plainnat}
\bibliography{amcs}

\end{document}